%% file: main.tex
\documentclass[lettersize,journal]{IEEEtran}
\usepackage{amsmath,amsfonts}
\usepackage{algorithmic}
\usepackage{algorithm}
\usepackage{array}
\usepackage{textcomp}
\usepackage{stfloats}
\usepackage{url}
\usepackage{verbatim}
\usepackage{graphicx}
\usepackage{cite}
\usepackage{subfiles}   
\usepackage{bm}         
\usepackage{bbm}        
\usepackage{makecell}   
\usepackage{xcolor}     
\usepackage{subcaption} 

\newcommand{\model}{\texttt}

\title{Deep COVID-19 Forecasting for Multiple States\\ with Data Augmentation}

\author{
  Chung Yan Fong and Dit-Yan Yeung
}

\begin{document}

\newcommand{\compilingmain}{yes}

\maketitle

\subfile{s0_abstract}

\subfile{s1_introduction}

\subfile{s2_related_works}

\subfile{s3_methodology}

\subfile{s4_da}

\subfile{s5_expts_n_results}

\subfile{s6_discussion}

\subfile{s7_conclusion}

\subfile{sL_references}

\end{document}

%% file: s0_abstract.tex
\begin{abstract}
In this work, we propose a deep learning approach to forecasting state-level COVID-19 trends of weekly cumulative death in the United States (US) and incident cases in Germany.  This approach includes a transformer model, an ensemble method, and a data augmentation technique for time series.  We arrange the inputs of the transformer in such a way that predictions for different states can attend to the trends of the others.  To overcome the issue of scarcity of training data for this COVID-19 pandemic, we have developed a novel data augmentation technique to generate useful data for training.  More importantly, the generated data can also be used for model validation. As such, it has a two-fold advantage: 1) more actual observations can be used for training, and 2) the model can be validated on data which has distribution closer to the expected situation.  Our model has achieved some of the best state-level results on the COVID-19 Forecast Hub for the US \cite{hub} and for Germany \cite{hubde}.
\end{abstract}

\begin{IEEEkeywords}
COVID-19 forecasting, time series prediction, transformer, data augmentation.
\end{IEEEkeywords}

\ifthenelse{\isundefined{\compilingmain}}{\subfile{sL_references}}{}

%% file: sL_references.tex
\bibliographystyle{IEEEtran}
\bibliography{IEEEabrv, ref}

%% file: s1_introduction.tex
\section{Introduction}
\label{sec:introduction}

COVID-19 pandemic has caused tremendous damages around the world.  Accurately predicting its trends, including the number of cases infected and deaths, remains one of the most essential tools for the public to understand the scale of the problem and plan for appropriate actions.

Researchers from all over the world have proposed a wide range of models to achieve this important task.  Compartmental models are one of the standard tools for modeling common epidemic dynamics.  They have also been widely used as the foundation of COVID-19 forecasting \cite{zou2020epidemic_ucla, li2020forecasting_delphi, gibson2020real_MechBayes, srivastava2020fast_alpha, arik2020interpretable_google}.  Based on some presumed mechanisms in epidemiology, these models describe how the population of different groups, such as the infectious and the healthy ones, affect each other and evolve in numbers.  During this pandemic, researchers have also proposed some other model variants to better capture particular situations that we have encountered in COVID-19,  such as considering untested and unreported cases \cite{zou2020epidemic_ucla}, non-pharmaceutical interventions (NPIs) \cite{li2020forecasting_delphi}, and population flows \cite{jia2020population_flow}.

Compared to the above approach, deep learning models, such as \cite{marzouk2021deep_egypt, yu2021covid_lstm, rodriguez2020deepcovid, jin2021interseries}, offer an alternative which does not require prior knowledge in epidemiology.  They are free from following some predefined modeling assumptions which are usually at best designers' approximations to the reality and hence may not be very accurate.  On the contrary, with a data-driven perspective, deep  learning models are trained to capture the underlying interactions of the real situation directly from the data collected.  They can easily handle various input types \cite{rodriguez2020deepcovid} and are able to discover more sophisticated and multivariate correlations \cite{jin2021interseries}.  However, the major drawback of this approach is that it generally requires a large amount of data for model training.  Since COVID-19 has a relatively short history, only with a few hundred days of observations recorded so far, training a deep learning model for generalized COVID-19 forecasting remains challenging.  For instance, models that learn only from past observations in Germany may find it difficult to infer the exceedingly high number of cases in November 2021.

Having to reserve part of the already scarce training data for validation is yet another issue.  Aiming for models that could generalize better in the situations to come, it is a common choice to reserve data of the most recent days for validation \cite{jin2021interseries}. Consequently, only earlier data can be used for training.  However, holding out the most recent part of data, arguably the most valuable part, from the training set will likely hinder the performance of the models.  Randomly sampling training examples for validation is neither a good choice, for COVID-19 has evolved and hence past observations may not be so indicative of what is going to happen in near future \cite{callaway2021beyond_nature}.  In summary, these common choices of validation splits may not be favorable in this situation and have room for improvement.

To overcome issues incurred by limited observed data, we propose a data augmentation (DA) method which can help generate synthetic data for either training or validation purpose, in particular for COVID-19 death and case forecasting.  With extra supply of validation data, the models can use the most recent data for training.  By using a transformer based deep neural network and a simple ensemble technique, our models, \model{HKUST-DNN} and \model{HKUST-DNN\_DA}, have achieved some of the best state-level forecasts among the models submitted to the COVID-19 Forecast Hub \cite{hub} and German COVID-19 Forecast Hub \cite{hubde} (Hubs).

The main contributions of our work are as follows:
\begin{itemize}
    \item We propose and examine a deep learning framework for the forecasting of state-level COVID-19 deaths and cases.
    \item Our data augmentation method can help generate synthetic data of COVID-19 deaths and cases.  Such data can be applied effectively for either training or validation purpose.
    \item Results of these methods have been submitted to the COVID-19 Forecast Hub for the US \cite{hub} and Germany \cite{hubde}.  We have achieved some of the best prediction results among many models proposed by different groups of researchers.
\end{itemize}

\ifthenelse{\isundefined{\compilingmain}}{\subfile{sL_references}}{}

%% file: s2_related_works.tex
\section{Related Works}
\label{sec:related-works}

\subsection{COVID-19 Forecast}

\subsubsection{Compartmental mechanistic models}
Compartmental mechanistic models are common tools for epidemic modeling \cite{weiss2013sir, zou2020epidemic_ucla, abbott2020estimating_delay, li2020forecasting_delphi, chen2020state_seir, srivastava2020fast_alpha, gibson2020real_MechBayes, karlen2020characterizing}.  In this domain, the whole population is divided into compartments that represent people in different conditions, such as the Susceptible, Infected and Removed in a standard SIR model \cite{weiss2013sir}.  By describing the dynamics between different groups, usually using a set of differential equations, these models can project into the future how populations of individual groups will evolve and hence infer the quantities of interest.

During this pandemic, researchers have proposed variants of compartmental models to take various dynamics into considerations, such as untested and unreported cases \cite{zou2020epidemic_ucla}, reporting delays \cite{abbott2020estimating_delay}, government interventions \cite{li2020forecasting_delphi}, and flows of travelers \cite{chen2020state_seir}.  Some of them proposed methods that can find model parameters in more effective ways \cite{srivastava2020fast_alpha}.

\subsubsection{Deep learning models}
Another popular approach to modeling the trend of this pandemic is by statistical learning using deep neural networks \cite{jin2021interseries, rodriguez2020deepcovid, yu2021covid_lstm, marzouk2021deep_egypt}, which generally formulated the tasks as time series prediction problems.  One advantage of this approach is that the models can easily handle various input types, including those with complex underlying interactions that are difficult to describe as in mechanistic models \cite{rodriguez2020deepcovid}.  They are also able to make use of sophisticated correlations, such as similarities across cities and time in the work of \cite{jin2021interseries}.  These models can generally be trained in an end-to-end manner without presuming any prior knowledge in epidemiology.  However, with such short history and limited data available for training, the models tend to overfit to previous observations.  Model training has to rely on various regularization schemes and the models used remain relatively shallow \cite{yu2021covid_lstm, marzouk2021deep_egypt, rodriguez2020deepcovid}.

\subsubsection{Hybrid and ensemble methods}
Some researchers also proposed to incorporate both techniques in their models.  While \cite{wu2021deepgleam} used a neural network model to predict the residual of a mechanistic model, \cite{arik2020interpretable_google} used a deep structure to predict the parameters of a complex compartmental model.

The COVID-19 Forecast Hub \cite{hub} and the German and Polish COVID-19 Forecast Hub \cite{hubde} have been established as platforms to collect and integrate forecasts from multiple groups of researchers using different methods.  They combine their submissions into a few ensemble models \cite{cramer2021evaluation_ensemble, bracher2021national_ensemble} to increase their stability and effectively reduce variance.

\subsection{Time Series Prediction with Deep Learning}

\subsubsection{Recurrent neural networks (RNN)}
Time series prediction is not a new application for deep neural networks.  Amazon has developed and launched a general framework based on auto-regressive RNN, named DeepAR \cite{salinas2020deepar}, for general time series probabilistic prediction.

\subsubsection{Graph neural networks (GNN)}
Another stream of work is to use variants of Graph Convolutional Neural Networks (GCRN) \cite{seo2018structured_GCRN} to better capture spatial-temporal correlation between multiple time series.  Many proposed methods have been tested on traffic datasets, predicting the traffic load on any spot of the road network.  While some methods work with a predefined graph \cite{li2018diffusion_DCRNN, yu2018spatio_STGCN}, other researchers have proposed methods which learn the underlying graph simultaneously during the training process \cite{bai2020adaptive_AGCRN, shang2021discrete_GTS}.

Although state-level pandemic forecasting is similar in nature to a traffic forecasting task, finding a graph which describes the inter-series spatial-temporal relationship is not as straightforward as in a road network.  Not many of these methods have been tested on COVID-19 prediction.  \cite{wu2021deepgleam} has incorporated a graph of travel data in Diffusion Convolutional Recurrent Neural Networks (DCRNN) \cite{li2018diffusion_DCRNN} to predict the residuals of a compartmental model.

\subsubsection{Transformers}
Since transformers \cite{vaswani2017attention} were proposed in 2017 demonstrating their effectiveness for many natural language processing (NLP) tasks, their adoption has subsequently been extended to other domains, such as the Vision Transformer (ViT) \cite{dosovitskiy2020image_ViT} in computer vision.  The multi-head self-attention mechanism of transformers has allowed every position of the input to find ways to relate to other positions effectively.

Researchers have also applied transformers to time series prediction \cite{li2019enhancing, zhou2020informer, tay2020long}, aiming to leverage the powerful self-attention mechanism to capture long-term dependency between the past and the future.  While they focused more on optimizing the model for longer sequence prediction tasks, such as reducing the quadratic complexity of self-attention, our work on the other hand aims to apply transformers in a different way such that multiple time series can interact with each other for relatively short-term prediction tasks.

\subsection{Data Augmentation for Time Series Data}

Data augmentation (DA) plays a crucial role in helping neural network models learn to be robust against legitimate variations.  It is a standard practice in computer vision applications \cite{bochkovskiy2020yolov4}.  For NLP tasks, researchers have also proposed a number of DA methods to augment the training data \cite{wei2019eda}.

There also exist some DA techniques for general time series data \cite{wen2020time_survey}.  Basic approaches include time-domain or frequency-domain transformations, while advanced techniques include modeling data statistically in the embedding space or using generative adversarial networks (GAN).  However, these techniques are not suitable for the COVID-19 forecast task, because neither of the operations proposed can reflect its epidemic nature.  The amount of training data currently available is very limited.

In this work, we propose to use compartmental mechanistic epidemic models to guide the augmentation of training data.  To the best of our knowledge, we are the first to synthesize data using this technique.

\ifthenelse{\isundefined{\compilingmain}}{\subfile{sL_references}}{}

%% file: s3_methodology.tex
\section{Forecasting Methodology}
\label{sec:methodology}

\subsection{Problem Definition}              
State-level COVID-19 forecasting can be formulated as a correlated time series prediction problem.  We are given a set of input time series $\bm{X}_i \in \mathbb{R}^{L{\times}d_{\mathit{in}}}$, where $i$ is the index for one of $N$ states or locations, $L$ denotes the number of time steps in each time series, and $d_{\mathit{in}}$ refers to the dimensionality of each observed event in time.  The goal is to predict the corresponding output time series at the $N$ locations, $\bm{Y}_i \in \mathbb{R}^{T{\times}d_{\mathit{out}}}$, where $T$ is the prediction horizon (number of time steps ahead) and $d_{\mathit{out}}$ is the output dimensionality.  One time step corresponds to one day.

To solve this problem, we seek to train a neural network which corresponds to a function $f_{\bm{\theta}}$ with a set of learnable parameters $\bm{\theta}$ such that the predictions made by the neural network aim to minimize some appropriate loss function $\mathcal{L}$,
i.e., $\min_{\bm{\theta}}\mathcal{L}\Big(
    \big[\bm{Y}_1, \bm{Y}_2, \ldots, \bm{Y}_N\big],
    f_{\bm{\theta}}\big(
        \big[\bm{X}_1, \bm{X}_2, \ldots, \bm{X}_N\big]
    \big)
\Big)$.

In the context of COVID-19 forecasting, the input features can include the number of confirmed cases, deaths, tests and vaccinations.  Other factors which are related to disease transmission in these locations can also be considered, such as social distancing policies and human mobility measures.  Empirically, we find that the four basic features of numbers of cumulative cases ($cum\_case$), cumulative deaths ($cum\_death$), weekly-incident cases ($inc\_case$) and weekly-incident deaths ($inc\_death$) generalize the best.  The results we are presenting in this paper use only these four basic features as inputs ($d_{in}=4$).  Such a reduced set of input features has also allowed us to develop a simple yet effective data augmentation technique to fuel the training process with more synthetic examples.

Regarding the outputs, our models are configured to predict multiple quantiles of a single target variable, i.e., the 23 quantile intervals required by the Hubs \cite{hub, hubde}.\footnote{\url{https://github.com/reichlab/covid19-forecast-hub/blob/master/data-processed/README.md#Data-formatting}}  This effectively allows our deep learning models to mimic probabilistic forecasts which may be essential in some downstream applications. Compared with more complicated probabilistic modeling techniques, quantile prediction oftens a much more efficient alternative \cite{cramer2021united}.  As such, a 23-dimensional output will be generated for each predicted target ($d_{out}=23$).

\subsection{Model Design}                    
We first apply a gated recurrent unit (GRU) layer to encode the $L$-day inputs of each location $\bm{X}_i$ into a latent representation.  Since the latent representations encode information of the individual time series separately, we refer to them as \textit{individually encoded representations}, $\bm{h}_i$.  An interaction layer is then introduced after the first encoding step to allow information to flow between time series.  Each time series can interact with every other time series to refine its own representation.  The result of such interaction is a set of \textit{collaboratively encoded representations}, $\bm{c}_i$.  Figure~\ref{fig:model} depicts the design of our model architecture which is conceptually simple. The following equations summarize the main computational steps:

\begin{figure*}[!h]
    \centering
    \subfloat[]{
        \includegraphics[width=5in]{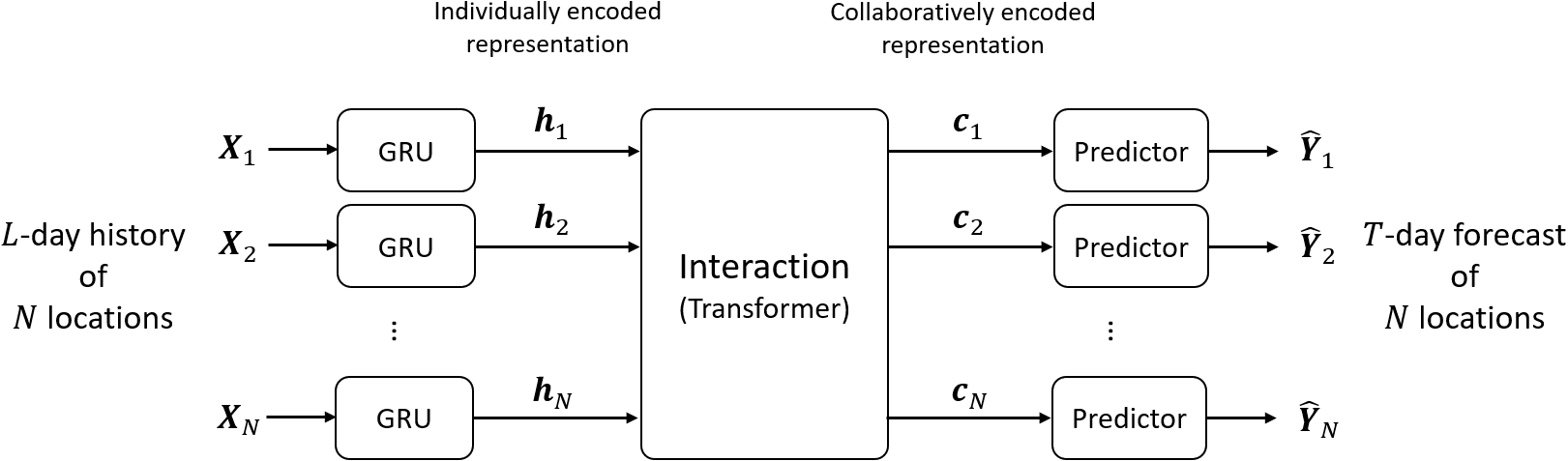}
    }
    \caption{Model architecture}
    \label{fig:model}
\end{figure*}

\begin{gather}
    \bm{h}_i = \text{GRU}\big(\bm{X}_i\big)
    \\
    \big[\bm{c}_1, \bm{c}_2, \ldots, \bm{c}_N\big] = \text{Transformer}\Big(\big[\bm{h}_1, \bm{h}_2, \ldots, \bm{h}_N\big]\Big)
    \\
    \hat{\bm{Y}}_i = \text{Predictor}\big(\bm{c}_i\big)
\end{gather}

We use a transformer encoder stack \cite{vaswani2017attention} as the medium of this interaction, for its multi-head self-attention mechanism allows information to be queried from one part of the inputs and be added to another part in the process.   Since our inputs $h_i$, which encode information of different geographical locations, do not have any specific sequential order, the original positional encoding layer of the transformer is simply omitted in our application.  After hyperparameter search, we decided to use a network configuration with a two-layer transformer encoder stack of 64-dimensional input/output and feed-forward layers, with 8 attention heads.

\subsection{Dual-residual Predictors}        
Instead of directly targeting the raw output values, our models are trained to predict some residuals which can be used to approximate the target values.  We propose a dual-residual estimation approach, which aims to combine two kinds of residuals, namely $\bm{R}^1$ and $\bm{R}^2$ residuals, for better prediction.  Equation~(\ref{eqt:r1r2}) depicts the predictor's formulation.

\begin{align}
    \big[\bm{R}^1_i, \bm{R}^2_i\big] &= \text{Linear}(\bm{c}_i)
    \nonumber
    \\
    \widehat{\bm{Y}}_i &= (1-\alpha)(\bm{Last\_observed\_value}_i+\bm{R}^1_i)
    \nonumber
    \\
    &\quad + \alpha(\bm{Projection}_i+\bm{R}^2_i)
    \label{eqt:r1r2}
\end{align}

\begin{figure}[!h]
    \centering
    \includegraphics[width=1.6in]{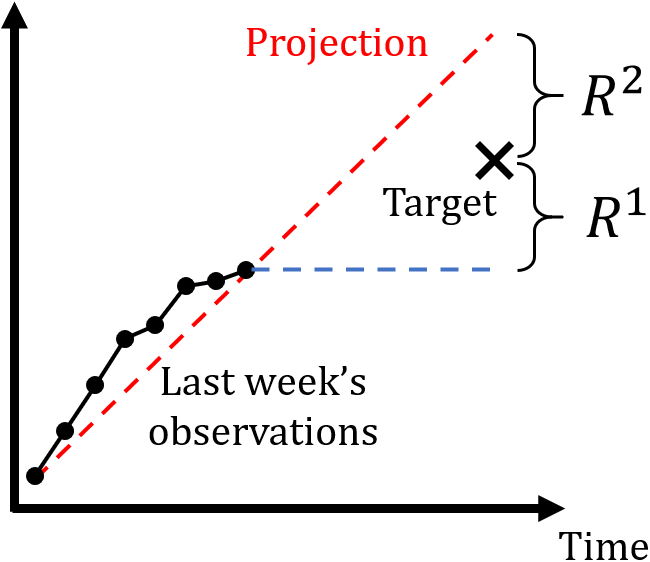}
    \caption{Illustration of the dual residuals $R^1$ and $R^2$.}
    \label{fig:r1r2}
\end{figure}

\subsubsection{$\bm{R}^1$ residuals}
They are the differences between the last observed values and the target values.  In some of our preliminary experiments, we found that predicting these differences is much more accurate than predicting the absolute target values.  In the case of forecasting the cumulative death counts, these differences are equivalent to the incident deaths since the last observation.

\subsubsection{$\bm{R}^2$ residuals}
On top of the $\bm{R}^1$ residuals, we also propose to incorporate the $\bm{R}^2$ residuals, which are the differences between the target values and a linear projection extrapolated from the previous week's observations.  If we see this projection as a baseline prediction, the $\bm{R}^2$ residuals measure how much the target values deviate from this baseline.

In Equation~(\ref{eqt:r1r2}), $\alpha \in [0,1]$ is a hyperparameter adjusting the weighting  between the two residuals to obtain a convex combination for predicting the target values. Separate predictors are used to approximate the two residuals.  Figure~\ref{fig:r1r2} shows an illustration of $\bm{R}^1$ and $\bm{R}^2$. 

\subsection{Loss Function}                   
\label{sec:loss-function}
It is usually more desirable to obtain probabilistic predictions in epidemic forecasting \cite{bracher2021evaluating_wis} because it gives us a better sense of the uncertainty of the forecast and, more importantly, allows us to prepare for the worst.  To approximate such outputs using a deterministic method, our model jointly forecasts a set of quantiles of the target variables.

In both the US and German Hubs \cite{hub, hubde}, their official ensembles are designed to aggregate models of 23-quantile forecast ($Q=23$).  Similar to previous work in quantile predictions \cite{takeuchi2006nonparametric, wen2017multi,rodrigues2020beyond}, we use a loss function which can produce multiple quantiles of the target variables, as shown in Equation~(\ref{eqt:pinball}) and (\ref{eqt:quantile_loss}).

\begin{gather}
    l_{\mathit{quantile}}(i,t,q)=
    \begin{cases}
        \tau_q(y_{i,t} - \hat{y}_{i,t,q})
        & \text{if \ } \hat{y}_{i,t,q} < y_{i,t}\\
        (1-\tau_q)(\hat{y}_{i,t,q} - y_{i,t})
        & \text{if \ } \hat{y}_{i,t,q} \geq y_{i,t}
    \end{cases}
    \label{eqt:pinball}
    \\
    \mathcal{L}_{\mathit{quantile}}=
    \frac{1}{N T Q}
    \sum_{i=1}^N
    \sum_{t=1}^T
    \Big{(}
    e^{t/\kappa}
    \sum_{q=1}^Q
    l_{\mathit{quantile}}(i,t,q)
    \Big{)}
    \label{eqt:quantile_loss}
\end{gather}

In Equation~(\ref{eqt:pinball}), $y_{i,t}$ represents the target value of the $i$-th location at time $t$, and $\hat{y}_{i,t,q}$ its predicted value at the $q$-th quantile.  While $q$ is the quantile index, $\tau_q$ denotes the $q$-th quantile value.  For example, when $Q=23$, the $12$-th quantile corresponds to the median, therefore $\tau_{12}=0.5$.

In order to put more emphasis on the forecast further into the future, we add an exponentially increasing term to put more weight on those errors.  The value of $\kappa$ is chosen such that the weight doubles after every seven days.

\begin{gather}
    \mathcal{L}_{\mathit{crossing}}=
    \frac{1}{N T Q}
    \sum_{i=1}^N
    \sum_{t=1}^T
    \sum_{q=2}^Q
    \text{ReLU}\big(\hat{y}_{i,t,q-1} - \hat{y}_{i,t,q}\big)
    \\
    \mathcal{L}=
    \mathcal{L}_{\mathit{quantile}}+
    \lambda_c\mathcal{L}_{\mathit{crossing}}
\end{gather}

Since individual quantiles are predicted separately, a major drawback of this approach is that some predicted values of the lower-quantiles may end up being larger than those of the higher-quantiles, which is known as the \textit{quantile crossing issue} \cite{takeuchi2006nonparametric, wen2017multi}.  To favor the monotonic property of the quantile values, we introduce an additional term $\mathcal{L}_{\mathit{crossing}}$ to regularize the output values.  This term penalizes errors proportionally to the magnitude of any quantile crossing found.  In our experiments, we set $\lambda_c=1$.

However, $\mathcal{L}_{\mathit{crossing}}$ alone is not sufficient to prevent all crossed quantile predictions and guarantee monotonicity at inference time.  To correct such occurrences, we also consider predictions at different quantiles as upper bounds for other quantiles below and lower bounds for higher ones.  For instance, if the $q$-percentile is predicted to be $\hat{y}$, percentiles below $q$ should also be capped below $\hat{y}$, while percentiles above should be at least $\hat{y}$.  From this perspective, in case crossing occurs, the predicted quantiles should effectively be swapped so as to satisfy the bounds that they are applying on each other.

\subsection{Model Selection and Ensemble}    
\label{sec:model-selection-and-ensemble}
One of the major challenges encountered in this study lies in the overfitting problem due to the limited training data.  Consequently, the performance of different randomly initialized models could result in a very high variance.  Single models are inadequate to achieve good generalization performance.  From some preliminary studies, we observed that low validation mean absolute errors (vMAE), as well as more epochs of training before early stopping based on the validation loss rebounds, are two good indicators for spotting the better forecasts.

\begin{gather}
    \label{eqt:weighted_sum}
    \hat{y}_{\mathit{ensemble}} = \frac{\sum_{k=1}^K w_k\hat{y}_k}{\sum_{k=1}^K w_k}
    \\
    w_k = \frac{1}{\text{vMAE}_k - \min_\ell \text{vMAE}_\ell /2}
    \nonumber
\end{gather}

By combining $K$ such models which rank best on both indicators, we can form an ensemble with better generalization performance. We have tried different weighting schemes to combine the outputs of the constituent models, such as the simple mean and exponentially weighted sum.  Empirically, we found that the weighted sum described in Equation~(\ref{eqt:weighted_sum}) achieved the best in general.  In this scheme, the contrast between the weights is boosted through reducing the denominator by half of their minimum.

Unless specified otherwise, each forecast we have made combines the best 10 models ($K=10$) out of five randomly seeded $\bm{R}^1$-only ($\alpha=0$), five $\bm{R}^2$-only ($\alpha=1$) and five balanced dual-residual ($\alpha=0.5$) models.

\ifthenelse{\isundefined{\compilingmain}}{\subfile{sL_references}}{}

%% file: s4_da.tex
\section{Data Augmentation}

\subsection{SIRD Data Model}
We propose to use a mechanistic compartmental model to augment actual observations with more synthetic data for training and validation.  We term this as our \textit{data model}.  This \textit{data model} can synthesize plausible trends of numbers of cases and deaths which could have happened in the past or happen in near future.  Such data are going to supplement the observed data when circumstances are never observed before.

Our \textit{data model} is based on standard compartmental mechanistic methods \cite{weiss2013sir, chen2020state_seir}.  It consists of four compartments, namely, the susceptible or uninfected ($S$), infectious ($I$), recovered ($R$), and death ($D$).  Their interactions are described by a set of difference equations shown in Equations~(\ref{eqt:sird1})--(\ref{eqt:sird4}).  In our setting, $S$ represents the uninfected population.  It can be approximated by the complement of cumulative confirmed case count, i.e. $S=1-cum\_case$, assuming a low reinfection rate \cite{hansen2021assessment_reinfection} and ignoring the unreported cases.  For simplicity, $D$ is the cumulative death count reported.  Therefore, they are both observed variables.

\begin{align}
    \label{eqt:sird1}
    S_{t+1} &= S_t -\beta{S_t}{I_t} \\
    \label{eqt:sird2}
    I_{t+1} &= I_t +\beta{S_t}{I_t} -\delta{I_t} -\gamma{I_t} +\omega{R_t}{I_t} \\
    \label{eqt:sird3}
    R_{t+1} &= R_t +\gamma{I_t} -\omega{R_t}{I_t} \\
    \label{eqt:sird4}
    D_{t+1} &= D_t +\delta{I_t}
\end{align}
Here $S$, $I$, $R$ and $D$ are normalized quantities in the range of 0 to 1 representing the corresponding populations. $I$ represents those who are infected and at the same time infectious to others, $R$ represents the people who have recovered from the disease but are still vulnerable to reinfection, and $\beta$, $\gamma$, $\omega$ and $\delta$ are parameters describing the rate of infection, recovery, reinfection and death, respectively.

\subsection{Model Fitting}

We want to find how $\beta$, $\gamma$, $\omega$ and $\delta$ have evolved in the past and recently.  If we can capture their distributions and sample from them, plausible instances of trends of numbers of cases and deaths can be generated.

To ease the fitting process, we introduce an extra learnable parameter $\eta$ to encode the initial conditions of $I$ and $R$, i.e., $I_{t_0}$ and $R_{t_0}$, where $t_0$ denotes the first timestep of every single fitting task.  It basically represents the portion of infected ones who are still infectious, rather than recovered.  By initializing $\eta$ to be close to $0.5$ in the beginning of every fitting, the optimization processes became more stable.

\begin{align}
    I_{t_0} &= (1-S_{t_0}-D_{t_0}) \eta \\
    R_{t_0} &= (1-S_{t_0}-D_{t_0}) (1-\eta)
\end{align}

Our fitting processes were done within some moving windows of lengths $W \in \{15, 16, ..., 28\}$.  These lengths are in number of days.  As a result, observations were cropped locally within 2 to 4 weeks and fitted to recover their characteristics during that particular period of time.  On a given day (timestep)~$t$, given a $W$-day history of cumulative deaths ($D_{t-W+1..t}$) and cumulative cases ($\mathbf{1}-S_{t-W+1..t}$), we solve for the parameters $\mathbf{\phi}(t,W) = [\hat{S}_{t_0}, \hat{D}_{t_0}, \beta, \gamma, \delta, \omega, \eta]$ which best describe the situations.  In this case, the first timestep $t_0=t-W+1$.

We designed a loss function $\mathcal{L}_{SIRD}$, as in Equation~(\ref{eqt:L_SIRD}), to minimize the sum of squared errors of the generated trends to the observed data. The parameters $\beta$, $\gamma$, $\delta$ and $\omega$ are regularized to be small and $\eta$ close to $0.5$.  $\lambda_D=0.5$, $\lambda_b=0.01$, $\lambda_o=0.03$ and $\lambda_h=0.001$ were the weights we used in our experiments.

\begin{multline}
  \label{eqt:L_SIRD}
  \mathcal{L}_{SIRD}=
  \frac{1}{W}
  \sum_{t'=t_0}^{t_0+W-1}
  \Big{(}
  (\frac{\hat{S}_{t'}-S_{t'}}{1-S_{t'}})^2
  + \lambda_D (\frac{\hat{D}_{t'}-D_{t'}}{D_{t'}})^2
  \Big{)}\\
  + \lambda_b (\beta^2 + \gamma^2 + \delta^2)
  + \lambda_o \omega^2
  + \lambda_h (\eta-0.5)^2
\end{multline}
The optimizations were computed by \verb|sklearn|'s \mbox{L-BFGS-B} algorithm \cite{byrd1995limited_LBFGSB}.  Section~\ref{sec:the-augmented-data} will show and discuss results of our fitted parameters.

For every location (state) on every day, there are results obtained using different window sizes.  This collection of fitted parameters allows us to capture a distribution of parameters, with which we will sample and generate synthetic data for DA.

\subsection{Data Generation}
\label{sec:data-generation}

We opt to approximate the fitted parameters by a multivariate Gaussian distribution $\Phi(t) = \mathcal{N}(\mathbf{\mu}_{\phi}(t), \mathbf{\Sigma}_{\phi}(t))$, where $\mathbf{\mu}_{\phi}(t)$ and $\mathbf{\Sigma}_{\phi}(t)$ are the mean and covariance matrix of $\mathbf{\phi}(t,W)$ across $W$.  By sampling instances of parameters from $\Phi(t)$, then inferring them using our \textit{data model}, new data instances of trends of cases and deaths can be generated at any given time $t$.

\subsubsection*{Projecting forward}
Since our SIRD \textit{data model} is simply a set of difference equations, with the intrinsic parameters $[\beta, \gamma, \delta, \omega, \eta]$ learned, it is able to infer longer sequences.  Presuming that the past trends continue for another $F$ days, we can approximate the following trend by inferring $\hat{S}_{t_1..t_2}$ and $\hat{D}_{t_1..t_2}$, where $t_1=t_0+F$ and $t_2=t_1+W-1$, from the model.  We will refer to this set of data as \textit{forward} in our later discussion.

\subsection{Data Splitting Schemes}

\begin{figure}[!h]
    \centering
    
    \begin{subfigure}[h]{0.5\textwidth}
        \centering
        \includegraphics[width=3.25in]{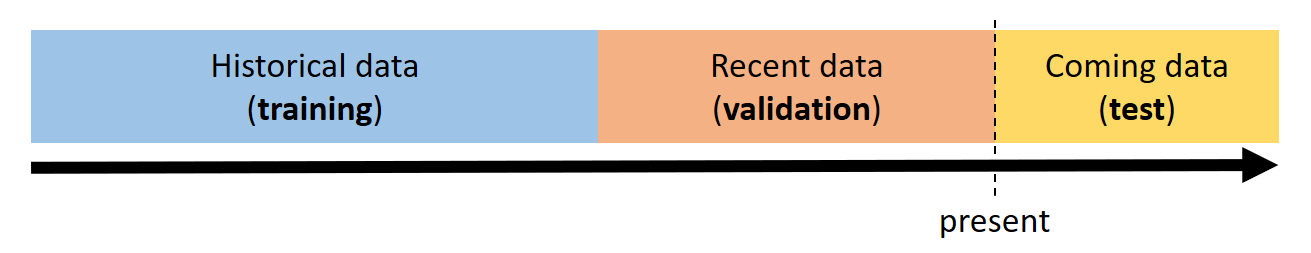}
        \caption{Usual data split.}
        \label{fig:usual-data-split}
    \end{subfigure}
    
    \begin{subfigure}[h]{0.5\textwidth}
        \centering
        \includegraphics[width=3.25in]{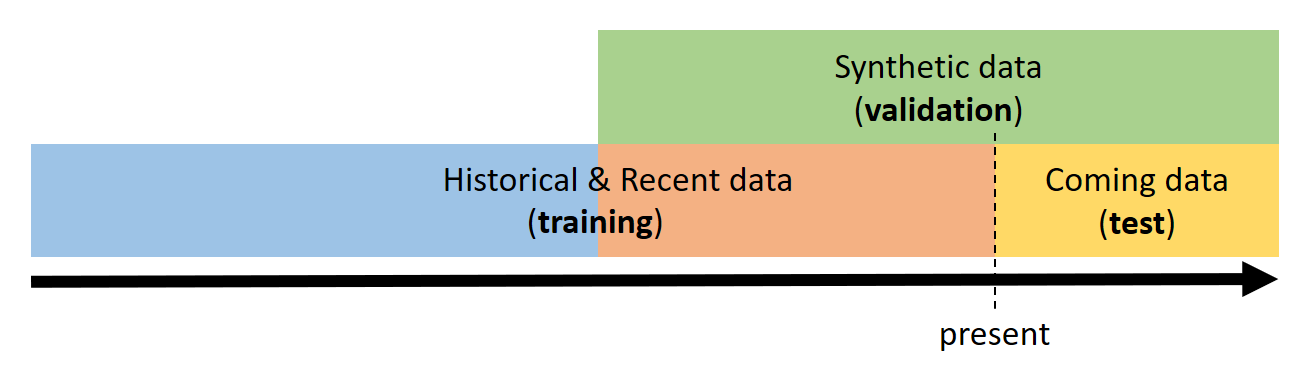}
        \caption{Data split utilizing synthetic data as validation.}
        \label{fig:da-val-split}
    \end{subfigure}

    \caption{Comparing data splitting schemes}
    \label{fig:data_split}
\end{figure}

A common choice of data split is to reserve the most recent data for validation purpose, as in Figure~\ref{fig:usual-data-split}.  This scheme assumes that the most recent data are close to the real test data.  However, the major drawback of this scheme is that a valuable subset of training data cannot be used to guide the minimization of the loss function to optimize the model's internal parameters.

In COVID-19 forecasting, this issue is particularly serious as the situations can change so quickly.  Our models would surely need the most recent data for training.  Another common choice is to randomly sample data for validation.  Yet, for the same reason, this is also not applicable to COVID-19 forecasting tasks.

With the synthetic data generated, we can use it for validation too, as shown in Figure~\ref{fig:da-val-split}.  Its advantage is two-fold.  On one hand, synthetic data with $F>0$ should be indicative of plausible outcome of the situations to come.  On the other hand, it releases the recent observations to be used as training data, thus allowing the forecasting model to be trained using the most recent data too.  Empirically, we can see that this is a better choice for sub-national Germany forecasts in our experimental results.

\ifthenelse{\isundefined{\compilingmain}}{\subfile{sL_references}}{}

%% file: s5_expts_n_results.tex
\section{Experiments and Results}
\label{sec:expts-n-results}

Since June 2021, we have started our experiments and begun to submit results of our \model{HKUST-DNN} model (trained on real data only, without DA) to the US Hub \cite{hub}, and, since November 2021, results of our \model{HKUST-DNN\_DA} model (trained on real data and validated by augmented data) to the German Hub \cite{hubde}.  Since then, based on the data collected up to every Saturday, our models are trained to predict $cum\_death$ in the US and $inc\_case$ in Germany for the four subsequent Saturdays, i.e., four different horizons spanning 1 to 4 weeks ahead.  Instead of predicting all the targets of the two countries, our scope of experiments has focused on predicting each country with a specific target, i.e., $cum\_death$ for the US and $inc\_case$ for Germany.

Our analysis below will focus on a 9-month window between June 2021 and February 2022, in which our models were designed and results were closely monitored.  This window covers the period when the dominating variant of  COVID-19 was transiting from Delta to earlier versions of Omicron.  For various issues and the performance after this period of time, we will discuss them later in Section~\ref{sec:discussion}.

\subsection{Evaluation Metrics}
In this paper, the forecast performance of different models will mainly be evaluated using the mean absolute error (MAE) of their point estimates (if absent, the estimated medians) and the weighted interval score (WIS) \cite{bracher2021evaluating_wis} of their distributional forecasts in a 23-quantile format.

\paragraph{Ground Truth}
\label{sec:ground-truth}
We adopt the death and case counts reported by Johns Hopkins University (JHU) \cite{jhu} and Robert Koch Institut (RKI) as ground truth values of the US and Germany, respectively.  The RKI's data is retrieved through the German Hub \cite{hubde}.  Please note that the JHU dataset is updated retrospectively, in which the latest values may be different from earlier versions.  As a matter of fact, there are frequently corrections made to the historical record of JHU's $cum\_death$.  To isolate errors induced by such corrections from the prediction errors, we opt to compare the predicted increments (since the day of the last observation) to the ground truth increments instead of comparing the actual predicted values to the corresponding ground truths directly.

\paragraph{Missing Data}
\label{sec:missing-data}
Every week, the  US Hub collects forecasts from dozens of models on various prediction targets.  Since there could be some missing forecasts (on certain dates or in some locations), we will only compare models which have contributed at least 90\% of the forecasts falling in our scope of interest, i.e., on $cum\_death$, 51 states (including Washington DC), 23 quantiles, 4 horizons (1- to 4-week), and within our 9-month window.  Furthermore, to maintain a fair comparison between the selected models which might still have slightly different missing values, our calculation of MAE and WIS will only include events in which all models have participated.  If there is any missing forecast from any model, that forecast will not be included into our calculation of MAE and WIS.  This same rule applies to the German forecasts.

\paragraph{Mean Absolute Error (MAE)}
For point estimates (or predicted medians), their performance is simply measured by their MAE to the ground truth.  However, since the reported state-level $cum\_death$ in the US can possibly be updated after our forecast, we opt to compare the predicted increments to the ground truth increments, instead of the absolute values, as explained above in Section~\ref{sec:ground-truth}.

\paragraph{Weighted Interval Score (WIS)}
For probabilistic forecast evaluation (in 23-quantile format), we adopt the WIS scoring method proposed by \cite{bracher2021evaluating_wis} to compare forecast submissions to the German Hub.  Unlike traditional methods such as the continuous ranked probability score (CRPS) for measuring the errors between continuous distributions, WIS only requires certain quantiles of such distributions to be evaluated, such as the 23-quantile format which is being used in this work.  It is also adopted by other researchers \cite{cramer2021evaluation_ensemble} to compare the results in  the US Hub.

\subsection{Results in the US (without DA)}
\label{sec:results-in-us}
Using our transformer-based deep learning model, we have participated in forecasting 1 to 4 weeks ahead (4 horizons) the $cum\_death$ in 51 states (including Washington DC) of the US since June 2021.  Our submitted results are named under \model{HKUST-DNN} in the US Hub \cite{hub}.

\begin{figure}[!h]
    \centering
    \includegraphics[width=3.5in]{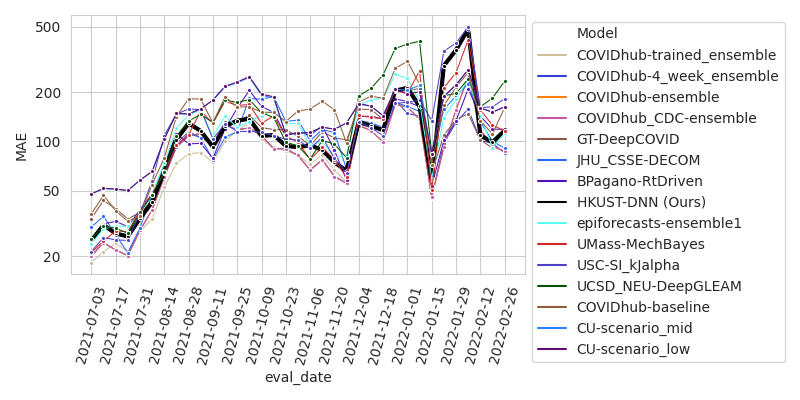}
    \includegraphics[width=3.5in]{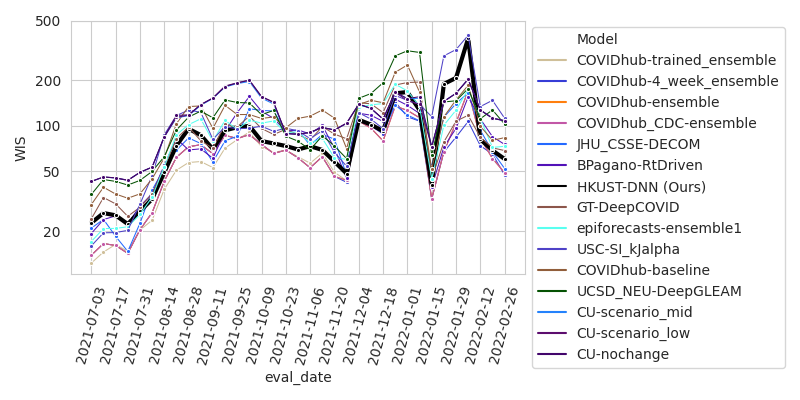}
    \caption{Trends of forecast performance (MAE and WIS) in the US}
    \label{fig:trend_of_mae_wis_us}
\end{figure}

Figure~\ref{fig:trend_of_mae_wis_us} shows the trends of MAE of the top 15 models in the US Hub based on the criteria mentioned in Section~\ref{sec:missing-data}.  Its horizontal axis represents the dates being forecasted, from July 2021 to February 2022.  The MAE of every forecasted date is an aggregated result of 4 different horizons (1 to 4 weeks ahead).

In terms of the $cum\_death$ forecast, our model \model{HKUST-DNN} has achieved a relatively low MAE compared to other models submitted to the Hub.  It also has comparable performance to the Hub ensembles named under "\model{COVIDhub...ensemble}".  Note that they are the combined and balanced results of all the models from various methods submitted to the Hub.


The WIS of forecasts from different models have very similar trends to their corresponding MAE.  It is also included here for completion.

Table~\ref{tab:mae_rank_wis_us} reveals the overall MAE over this period of time.  They are the aggregated results from all 51 states and 4 horizons (1 to 4 weeks ahead).  For each date, horizon and state, each model has a rank among others, depending on who has a lower MAE.  The average of these ranks is used to sort the entries in the table in ascending order.  In term of this average rank, our model \model{HKUST-DNN} has ranked relatively high among others.  Note that the top-5 models are all Hub ensembles which are aggregated results of all submitted models.



\subfile{table/tab_us_mae-rank-wis}

\subsection{Results in Germany (with DA)}
\label{sec:results-in-germany}
We have also tested our model with DA on forecasting the sub-national $inc\_case$ of the 16 states in Germany.  Adopting the naming convention of the Germany Hub \cite{hubde}, these 16 German states are also aliased as GM01 to GM16.\footnote{\url{https://github.com/cfong32/covid19-forecast-hub-de/blob/master/template/state_codes_germany.csv}}

Compared to the US, Germany has a much smaller dataset, attributing to its fewer states (also with a smaller population and territory).  From our observation, the trends of $case$ and $death$ in Germany exhibit less variation, especially prior to the recent waves of Omicron variants.

Therefore, in order to avoid overfitting to the small training dataset, we introduce our DA method in training the models for Germany.  Empirically, we found that the models trained with our DA method perform better than those without.  Based on the following divergence analysis, we will also show that our augmented data has a distribution closer to the test data (to be forecasted), rather than the real training data (historical data).

\subsubsection{Augmented Data}
\label{sec:the-augmented-data}

\paragraph{Fitted parameters}

\begin{figure}[!h]
    \centering
    \includegraphics[height=1.7in]{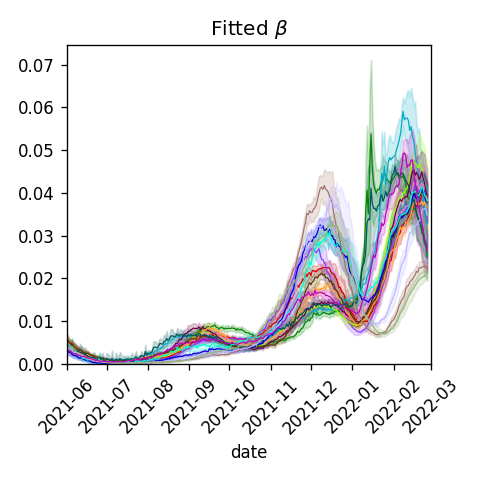}
    \includegraphics[height=1.7in]{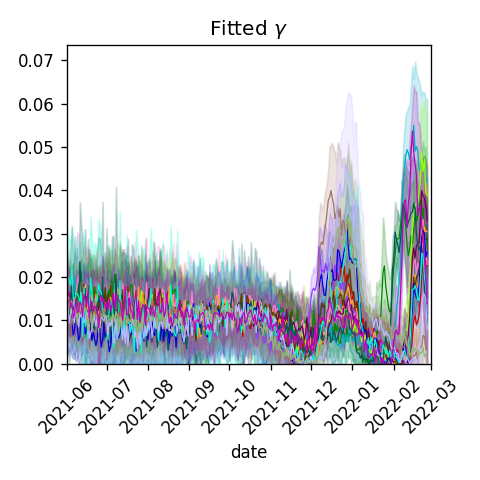}
    \includegraphics[height=1.7in]{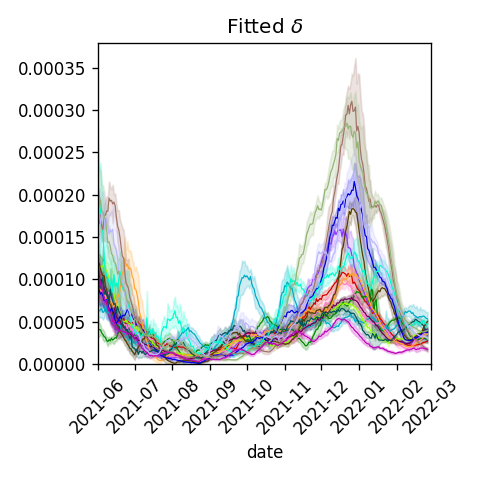}
    \includegraphics[height=1.7in]{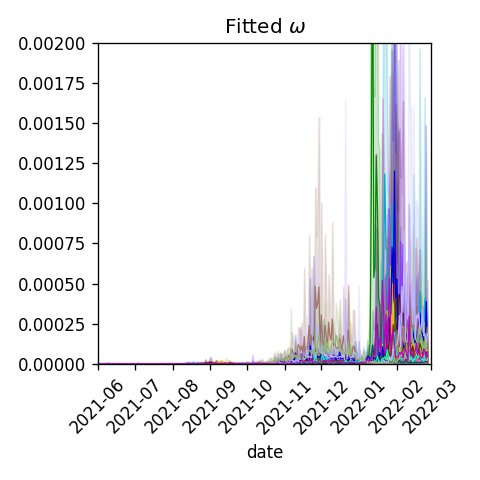}
    \includegraphics[height=1.7in]{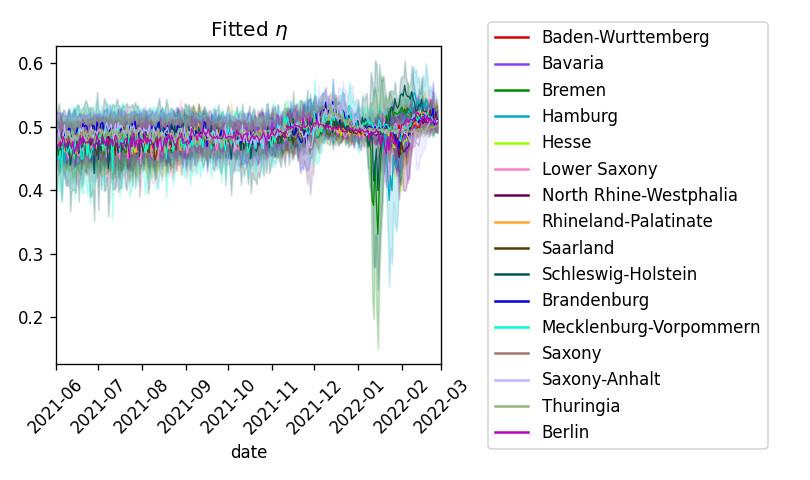}
    \caption{Fitted parameters of the 16 states in Germany.}
    \label{fig:fitted_params}
\end{figure}

Figure~\ref{fig:fitted_params} shows the results of the fitted parameters of our \textit{data model}.  On every day (or timestep) $t$, the parameters are obtained by fitting our SIRD model to a recent history of $W \in\{15,16,\dots,28\}$ days in each of the 16 states.  The parameters $\beta$, $\gamma$, $\omega$ and $\delta$ depict the rates of infection, recovery, reinfection and death, respectively, $\eta$ is the initial portion of infectious population among the infected ones, and the ratio of $\eta:1-\eta$ represents the proposed initial ratio of $I$ to $R$.  Since $\eta$ is initialized to 0.5 before fitting, unless the \mbox{L-BFGS-B} optimization needs a much lower or higher value to explain the trends, it tends to converge to local optima around 0.5.

\paragraph{Generating new data}
Sampling from the observed or fitted distributions of $S$, $D$, $\beta$, $\gamma$, $\omega$, $\delta$ and $\eta$, we obtained new parameters to generate synthetic trends for DA.  This process was described in Section~\ref{sec:data-generation}.

\begin{figure}[!h]
    \centering
    \includegraphics[width=3.5in]{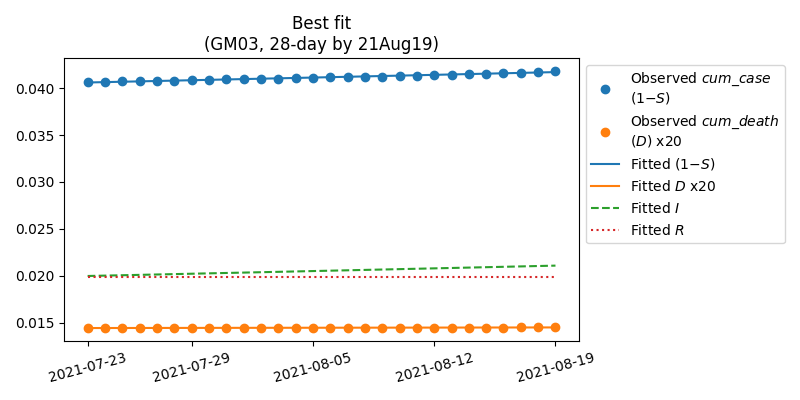}
    \includegraphics[width=3.5in]{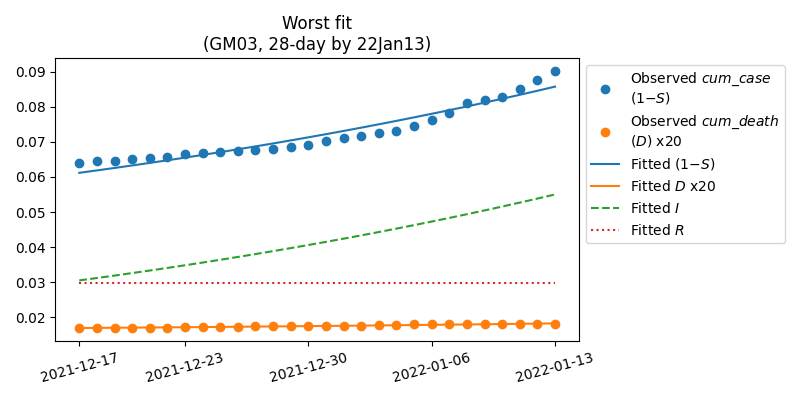}
    \caption{Best- and worst-fitted state-level trends in Germany}
    \label{fig:worst-fits}
\end{figure}

Figure~\ref{fig:worst-fits} shows samples of the best and the worst fitted 28-day trends ($W=28$).  They respectively represent the most easy and difficult cases that our SIRD model has comprehended so far.  These examples are shown primarily to demonstrate the quality of the fitting processes.

A couple of points are worth noting.  First of all, for visualization purpose, the death counts are multiplied by a factor of 20, so that their trends can be visually compared more easily.  Green and red dashed lines are the inferred $I$ and $R$.  Secondly, unlikely the real observations which are more noisy, these fitted trends are smoother and lack of noise.  Yet we hope that they can still capture much information about the situations that are plausible to happen.

\paragraph{Divergence analysis}
\label{sec:divergence-analysis}

To verify the suitability of our synthetic data, we conducted a divergence analysis between the original training data and our synthetic data.  In particular, we want to see if the synthetic data is close to the test data, so as to enhance the predictability over the situations in the near future.

\begin{figure}[!h]
    \centering
    \includegraphics[width=3.5in]{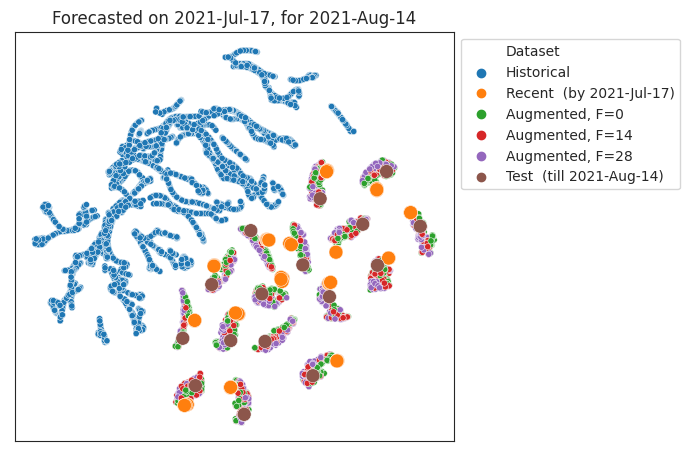}
    \includegraphics[width=3.5in]{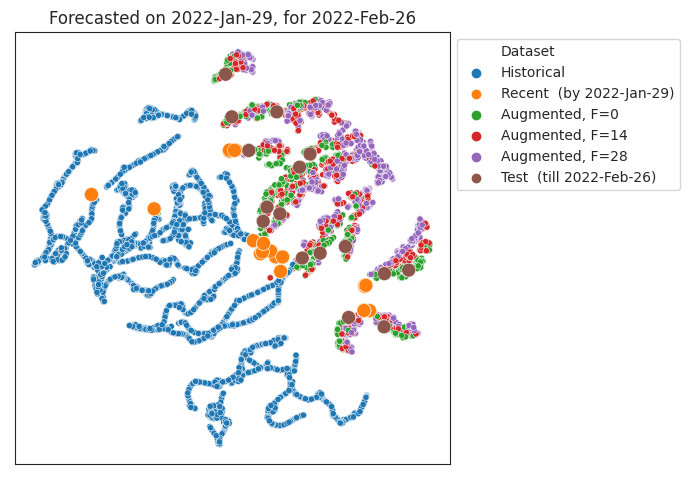}
    \caption{t-SNE plots of datasets on 2021-Jul-17 and 2022-Jan-29}
    \label{fig:tsne}
\end{figure}

We first visualize the data using t-SNE \cite{van2008visualizing_tsne} plots.  Training examples $[X_i, Y_i]$ are first serialized into vectors $Z_i \in \mathbb{R}^{L{\times}d_{\mathit{in}}+T{\times}d_{\mathit{out}}}$.  Figure~\ref{fig:tsne} shows two t-SNE plots which visualize these vectors in a 2-D space.  Every plot corresponds to a particular date of forecast.  In order to show the test and recent observations more clearly, their point sizes are enlarged.

\begin{figure}[!h]
    \centering
    \includegraphics[width=3.5in]{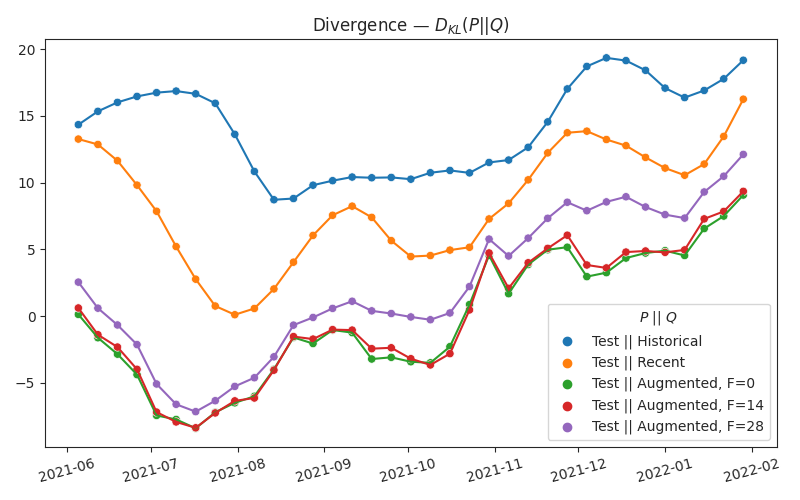}
    \caption{Trend of the estimated divergence}
    \label{fig:divergence}
\end{figure}

Two particular dates are shown.  The first plot in Figure~\ref{fig:tsne} visualizes various available datasets when forecast was carried out on 2021-Jul-17, with $L=14$ and $T=28$.  Its test data refers to a four-week horizon till 2021-Aug-14.  The recent observations ($L+T+7$ days by 2021-Jul-17) were supposed to form the validation set in the usual practice without DA.  All historical data (approximately two years) was available for training.  All synthetic datasets (with $F \in {0, 14, 28}$) were generated based on observations as of Jul-17.  Based on our divergence analysis, as shown in Figure~\ref{fig:divergence}, this time reached the lowest divergence so far between the test data and our generated data.  For comparison, the latter plot in Figure~\ref{fig:tsne} depicts the circumstance on 2022-Jan-29.  In contrast, It has the highest divergence so far in our experiments.

Yet, both plots of different dates visually demonstrate that, compared to the available training data (historical as well as recent observations), our generated data could produce a smaller divergence from the tests.  As the pandemic has evolved so much since its emergence in 2019, its historical course of observations (real training data) may become limited in capturing the distribution of trends which we are trying to predict (test cases).

Similar to the idea of \cite{wang2009divergence_kNN}, we estimate the KL divergence using a $k$-nearest-neighbor (kNN) density estimator.  In Equation~\ref{eqt:DKL_2}, $z_i$ denotes every sample drawn from distribution $P$, $v_i$ is the spherical volume of its $k$ nearest neighbors occupied in $Q$, and $n$ and $m$ are the numbers of samples obtained from $P$ and $Q$, respectively.

\begin{align}
    D_{KL}(P||Q)
    &= \mathbb{E} \Big{(} P(Z) log \frac {P(Z)} {Q(Z)} \Big{)} \label{eqt:DKL_1} \\
    &\approx \frac{1}{n} \sum_{i=1}^{n}
        log \frac {P(z_i)} {Q(z_i)} \label{eqt:DKL_2}
    \\
    &\approx \frac{1}{n} \sum_{i=1}^{n}
        log \frac {1 / n} {k / m v_i} \label{eqt:DKL_3}
\end{align}

In our analysis, $P$ corresponds to the distribution of the test data, which is the target our models need to capture.  Since every Saturday corresponds to a new forecast with an updated model trained on the latest data, the test data for each model normally contains just a few samples (i.e., one sample from each state $i$), which are hardly formulated by a density function.  We sought to approximate it by a uniform weighting across its samples; therefore, $P(z_i)$ is simplified to $1/n$ in Equation~(\ref{eqt:DKL_3}), whereas $Q(z_i)=k/mv_i$ is still estimated by its $k$-nearest-neighbors density.

Figure~\ref{fig:divergence} shows a comparison in terms of the estimated divergence, from the test data to the 1) historical data (training set), 2) recent data (usual validation set), and 3) our augmented data.  The densities of $Q$ are approximated by the third nearest neighbor of the data points, i.e., $k=3$.  From the comparison, we can observe that the original training data has a relatively larger divergence from the test set.  We can see that our synthetic data will likely be helpful for prediction because it has a smaller divergence to the target data.

Please note that the negative values in the estimated divergence suggests that our kNN approximation may produce a density $Q(z_i) > 1/n$.  It means that the third nearest neighbor may be so close to the query point (test data) that it over-estimates the density around.  Yet it is still a fair comparison, because all datasets are estimated using the same set of $P(z_i)$.  

As a result, it shows that, over the whole period of time, the generated data are closer to the test data, rather than the training or the usual validation (recent) data.  Consequently, it is expected that the generated data can be a good supplement as an additional supply of training data.

\subsubsection{Forecast Performance in Germany}

\paragraph{Models included}

Our results submitted to the German Hub are named \model{HKUST-DNN\_DA}, which is trained on the real data and validated using synthetic data.  It will also be aliased as \model{HKUST-DNN\_DA-val} in our following discussion.  We have also included \model{HKUST-DNN} and \model{HKUST-DNN\_DA-train} in this paper for comparison.  They represent our models trained without DA and using DA as training data, respectively.

\begin{figure}[!h]
    \centering
    \includegraphics[width=3in]{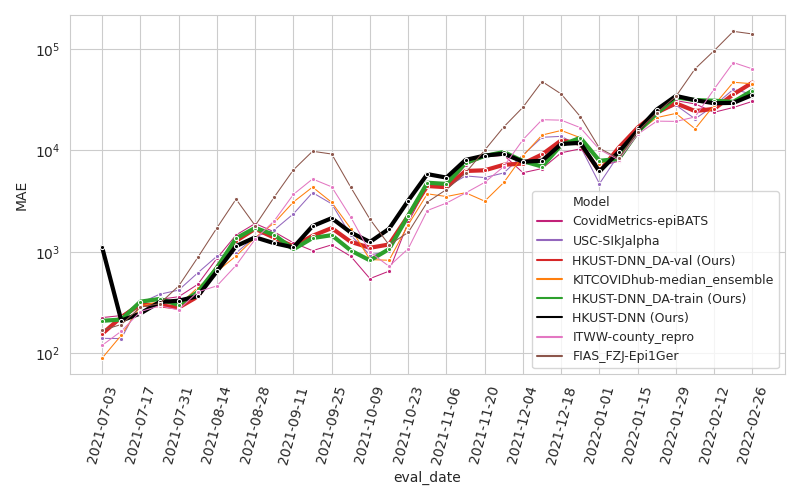}
    \includegraphics[width=3in]{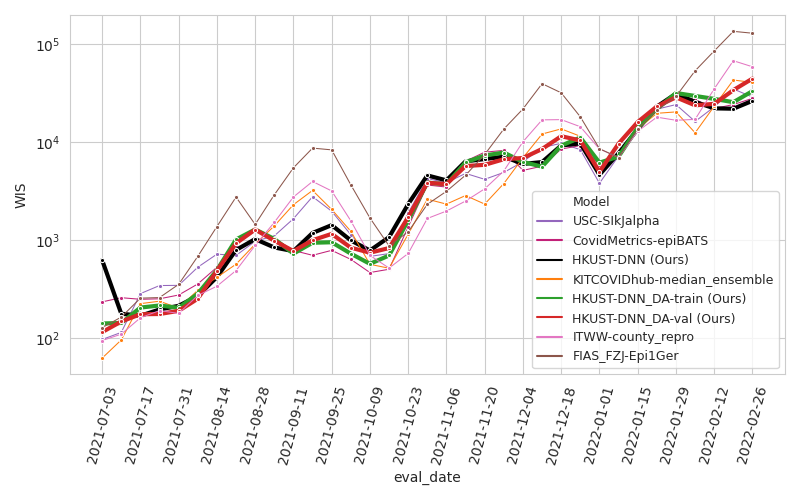}
    \caption{Trends of forecast performance (MAE and WIS) in Germany}
    \label{fig:trend_of_mae_wis_de}
\end{figure}

Figure~\ref{fig:trend_of_mae_wis_de} shows a comparison of three of our trained models to other models available in the German Hub.  \model{HKUST-DNN} corresponds to our model without DA (trained only on historical observations) and \model{HKUST-DNN\_DA-val} is our model trained using DA for validation.  It is equivalent to our model \model{HKUST-DNN\_DA} submitted to the German Hub.  \model{HKUST-DNN\_DA-train} is our model trained using DA as part of the training set, keeping the recent observations for validation.  Please note that the results of \model{HKUST-DNN} and \model{HKUST-DNN\_DA-train} are only available in our GitHub fork\footnote{\url{https://github.com/cfong32/covid19-forecast-hub-de}} of the German Hub.

Here, augmented data are generated using a 14-day forward ($F=14$) setting.  To supply data which is closer to the situation to be forecasted, only new synthetic data which generated in previous seven days were included in each training or validation.

Figure~\ref{fig:trend_of_mae_wis_de} shows the overall performance of our models compared to other models submitted to the German Hub.  They are measured in MAE and WIS, same as for the US.


The models being compared are submissions from other research groups, such as 
FIAS\_FZJ-Epi1Ger from \cite{barbarossa2020impact_hubde_FIAS}  
and USC-SIkJalpha from \cite{srivastava2020fast_alpha}.  
For details of these models, please refer to their published meta data on the Hub \cite{hubde}.  To the best of our knowledge, ours are the only deep learning models based on neural networks.

One of the Hub ensembles (KITCOVIDhub-median\_ensemble) is also compared.\footnote{Since the German Hub ceased publishing official KITCOVIDhub-median\_ensemble forecasts after 2021-Dec-06, we have continued calculating it for subsequent dates.}  It is a simple quantile-wise median of all the submitted models, including our \model{HKUST-DNN\_DA}.

Comparing our own results, both our \model{HKUST-DNN\_DA-val} (red) and \model{HKUST-DNN\_DA-train} (green) models could result in lower errors than the one without DA (black).  It shows that our augmented data could help our deep learning models to capture and predict the trends.  Especially on Jul-03, \model{HKUST-DNN} resulted in a relatively large error due to overfitting to training data.  With the help of DA, in either validation or training, our models resulted in a much lower error.

Compared with some other models in the Hub, especially during the plateaus in 2021 Sep and Dec, our model \model{HKUST-DNN\_DA-val} was able to deliver more stable performance.

Table~\ref{tab:mae_rank_wis_de} shows the overall MAE, WIS and rank of all the models.  In terms of MAE, \model{HKUST-DNN\_DA-val} has achieved the second best among all submissions.  The average rank is calculated similarly to the US results.

\subfile{table/tab_de_mae-rank-wis}

\subsection{Our DA on Other Deep Learning Models}
\label{sec:da-on-other-models}

\subsubsection{Other Models}
\label{sec:other-models}
As a proof of effectiveness, we have tried to apply our generated data on two other deep learning models.  One of them is based on a simple LSTM encoder-decoder architecture.  Each of the encoder and decoder is simply a 2-layer 64-unit LSTM.  Another model is the Adaptive Graph Convolutional Recurrent Network (AGCRN) \cite{bai2020adaptive_AGCRN}.  It is a multivariate time series prediction architecture proposed for traffic forecasting in road networks.  By applying the same quantile loss as described in Section~\ref{sec:loss-function}, on these two architectures, they are also able to produce 23-quantile outputs for each of the 16 states in Germany.

\subsubsection{Results}
\label{sec:other-models-results}
Table~\ref{tab:da-hkust-agcrn-lstm} shows a comparison of our \model{HKUST-DNN} model with the two baselines with and without DA, during the period from 2021-Jun-01 to 2022-Jan-01.  With our DA used for validation (\model{DA-val}) and saving the recent data for training, these models are all improved with an overall reduction in MAE.

As the pandemic progresses, the number of cases sometimes can vary by orders of magnitude.  The overall MAE or WIS could be dominant by times when the number of cases is high.  Therefore, on top of the total reduction in percentage, here we also calculate the per-date reduction which measures the average of percentage reduction on different target dates.  In terms of this per-date MAE and WIS metrics, the performance of our model \model{HKUST-DNN} and \model{AGCRN} can aslo be boosted by the augmented data.

However, the WIS evaluation seems to suggest that the improvement in  probabilistic forecasts (in 23-quantile format) is not as prominent as in the point (median) forecasts.  The smooth and simple generated data was not able to promote performance for simple models like the standard \model{LSTM} architectures.

\begin{table}[!h]
    \centering
    \caption{MAE and WIS reduction using DA
    for validation}
    \label{tab:da-hkust-agcrn-lstm}
    \begin{tabular}{lrrrr}
    \hline
    \multicolumn{5}{c}{MAE} \\
    \hline
                model &  no-DA & DA-val &  Total reduction &  Per-date reduction \\ \hline
    \model{HKUST-DNN} & 3726.1 & 3349.6 &           10.1\% &              11.3\% \\ \hline
    \model{AGCRN}     & 3447.0 & 3298.7 &            4.3\% &               3.5\% \\ \hline
    \model{LSTM}      & 3412.2 & 3337.4 &            2.2\% &              -2.6\% \\

    \\
    \hline
    \multicolumn{5}{c}{WIS} \\
    \hline
                model &  no-DA & DA-val &  Total reduction &  Per-date reduction \\ \hline
    \model{HKUST-DNN} & 2836.1 & 2901.6 &           -2.3\% &               6.1\% \\ \hline
    \model{AGCRN}     & 2770.7 & 2633.3 &            5.0\% &               4.5\% \\ \hline
    \model{LSTM}      & 2669.5 & 3138.8 &          -17.6\% &             -14.4\% \\ \hline
    
    \end{tabular}
\end{table}

\subsection{Ablation Study on Dual-residual}
\label{sec:ablation-study-on-dual-residual}

To show the benefit of using our proposed dual-residual mechanism, we have carried out the following ablation study.

Table~\ref{tab:different-residual} shows a comparison of our submitted models \model{HKUST-DNN} and \model{HKUST-DNN\_DA} with three other small ensemble models under different residual settings ($\alpha \in \{0, 0.5, 1\}$).  The suffixes \model{\_R1}, \model{\_R2} and \model{\_dualR} indicate the dual-residual settings of $\bm{R}^1$-only ($\alpha=0$), $\bm{R}^2$-only ($\alpha=1$) and balanced dual-residual ($\alpha=0.5$), respectively.  These three residual-specific models are in fact smaller ensembles combining five composites of different random seeds, while the submitted ones are bigger ensembles combining the best 10 of all these 15 composites.  Please refer to Section~\ref{sec:model-selection-and-ensemble} for more details of the selection process for the best.

To help focus on this comparison, the reported average ranks and standard deviation of ranks are calculated among these four models only.

As \model{HKUST-DNN} and \model{HKUST-DNN\_DA} are ensembles of the other models, no wonder their average and variance of ranks are the lowest.  Nonetheless, this comparison clearly shows that balanced dual-residual has performed more steadily (with low rank SD) and is on average better (with lower average rank) than either one of the $\bm{R}^1$-only or $\bm{R}^2$-only setting.

\begin{table}[!h]
    \centering
    \caption{Comparison of difference residuals}
    \label{tab:different-residual}
    \begin{tabular}{lrrrr}
        \hline
        \multicolumn{5}{c}{$cum\_death$ forecast in the US} \\
        \hline
                       Model &  Avg. rank↓ &  Rank SD &    MAE &    WIS \\ \hline
        
           \model{HKUST-DNN} &        2.43 &     0.83 & 147.3 & 109.6 \\ \hline
    \model{HKUST-DNN\_dualR} &        2.47 &     1.06 & 142.6 & 103.5 \\ \hline
       \model{HKUST-DNN\_R1} &        2.51 &     1.22 & 169.7 & 124.8 \\ \hline
       \model{HKUST-DNN\_R2} &        2.59 &     1.30 & 139.3 & 107.3 \\


        \\
        \hline
        \multicolumn{5}{c}{$inc\_case$ forecast in Germany} \\
        \hline
                       Model &  Avg. rank↓ &  Rank SD &     MAE &     WIS \\ \hline
    
       \model{HKUST-DNN\_DA} &        2.30 &     0.63 & 8555.8 & 7974.0 \\ \hline
\model{HKUST-DNN\_DA\_dualR} &        2.44 &     0.87 & 7909.3 & 7357.6 \\ \hline
   \model{HKUST-DNN\_DA\_R2} &        2.51 &     1.37 & 9000.9 & 8480.4 \\ \hline
   \model{HKUST-DNN\_DA\_R1} &        2.74 &     1.37 & 8529.5 & 7597.1 \\ \hline
    

    \end{tabular}
\end{table}

\ifthenelse{\isundefined{\compilingmain}}{\subfile{sL_references}}{}

%% file: table/tab_us_mae-rank-wis.tex
\begin{table}[!ht]

\centering
\caption{Overall performance of {$cum\_death$} forecast in the US}
\label{tab:mae_rank_wis_us}

\begin{tabular}{lrrr}
\hline
                           Model &  Avg. rank↓ &     MAE &     WIS \\ \hline

      COVIDhub-trained\_ensemble &        9.26 &   83.07 &   61.17 \\ \hline
      COVIDhub-4\_week\_ensemble &        9.41 &   84.80 &   61.72 \\ \hline
               COVIDhub-ensemble &        9.42 &   86.63 &   63.45 \\ \hline
          COVIDhub\_CDC-ensemble &        9.42 &   86.63 &   63.45 \\ \hline
  KITmetricslab-select\_ensemble &       10.23 &  376.83 &  358.43 \\ \hline
  
                 JHU\_CSSE-DECOM &       10.74 &  102.01 &   74.77 \\ \hline
                HKUST-DNN (Ours) &       10.88 &  107.84 &   81.45 \\ \hline
                 USC-SI\_kJalpha &       11.22 &  116.46 &   97.48 \\ \hline
    SteveMcConnell-CovidComplete &       11.25 &  763.33 &  754.13 \\ \hline
          epiforecasts-ensemble1 &       11.71 &  113.51 &   84.43 \\ \hline
          
                    GT-DeepCOVID &       11.94 &  101.72 &   83.89 \\ \hline
                BPagano-RtDriven &       12.13 &  103.21 &   80.00 \\ \hline
              MIT\_CritData-GBCF &       12.54 &  147.06 &  129.45 \\ \hline
             UCSD\_NEU-DeepGLEAM &       13.14 &  133.26 &  109.72 \\ \hline
               COVIDhub-baseline &       13.41 &  134.83 &  102.03 \\ \hline
               

\end{tabular}

\end{table}

%% file: table/tab_de_mae-rank-wis.tex
\begin{table}[!ht]

\centering
\caption{Overall performance of {$inc\_case$} forecast in Germany}
\label{tab:mae_rank_wis_de}

\begin{tabular}{lrrr}
\hline
                        Model &  Avg. rank↓ &      MAE &      WIS \\ \hline

 KITCOVIDhub-median\_ensemble &        3.93 &  8584.38 &  7308.00 \\ \hline
           ITWW-county\_repro &        4.11 & 10849.09 &  9355.58 \\ \hline
         CovidMetrics-epiBATS &        4.39 &  7698.64 &  6993.11 \\ \hline
     HKUST-DNN\_DA-val (Ours) &        4.42 &  8556.37 &  7978.05 \\ \hline
   HKUST-DNN\_DA-train (Ours) &        4.50 &  8719.78 &  7571.46 \\ \hline
                USC-SIkJalpha &        4.52 &  8445.18 &  6894.39 \\ \hline
             HKUST-DNN (Ours) &        4.89 &  8789.48 &  7019.20 \\ \hline
            FIAS\_FZJ-Epi1Ger &        5.24 & 21491.14 & 18783.82 \\ \hline

\end{tabular}

\end{table}

%% file: s6_discussion.tex
\section{Discussion}
\label{sec:discussion}

\subsection{Probabilistic Forecast Performance}

When the available training data is limited, we need to generate some augmented data for training or validation so that the models will not overfit the limited dataset.  In the case of COVID-19 forecasting, we have shown that our DA approach was effective to avoid such overfitting problem.  Our synthetic data allows the models to consider situations which have not happened before, and thus be more generalizable to other plausible trends.

However, we have observed that the DA process adopted somehow hurts the probabilistic forecast performance, which is evident in limited improvement in WIS.  In our experiments, we have often observed that the faster the trends rose or fell, the more certain and less dispersed the prediction would be.  This phenomenon could be quite counter-intuitive, because normally when changes come sudden, uncertainty should increase.  The degradation in probabilistic forecast performance and WIS could possibly be attributed to the simplicity of our data model.  Compared to real data, our generated data is indeed lack of variety and noise.  Incorporating some noise into the data model would be one of the directions to explore in our future work.

\subsection{Our Data Model}

Our SIRD data model is relatively general and simple.  It is enough to provide an easy access to the trends of some latent variables like the rates of infection, recovery, reinfection and death, i.e., $\beta$, $\gamma$, $\omega$ and $\delta$, respectively.  However, our way of fitting the curves window by window has violated the continuity of some continuous values of $S$, $I$, $R$ and $D$.  This could cause certain errors in our estimation.  However, since the whole trends are guided by the observed $S$ and $D$, the overall error should be limited.

Due to the simplicity of our data model, it may not be able to capture complicated interactions between recent variants, such as the Omicrons.  Unfortunately, especially in our recent forecasts, we observe that the DA setting is not straightly better than the no-DA setting.  Using real data for validation could sometimes performs better than synthetic data.  This could mean that the parameters captured by our model cannot reflect the rapidly evolving situations.  To cope with this, we may need to explore more sophisticated data models or ensemble schemes.  This requires more detailed studies which will be left for our future work.
    
\ifthenelse{\isundefined{\compilingmain}}{\subfile{sL_references}}{}

%% file: s7_conclusion.tex
\section{Conclusion}
\label{sec:conclusion}
In this work, we have proposed a deep learning method for COVID-19 forecasting.  The whole method consists of a transformer model, an ensemble method together with a data augmentation method.  While the transformer model leverages signals from the trends of multiple states, the ensemble alleviates the problem of high prediction variance.  Together, they have overcome some problems induced by the limited training data and have achieved some of the best results of cumulative death prediction in the US Hub \cite{hub}.

Due to the limit training data available for training and validation, our DA method has been shown to be critical in enabling our German sub-national level case prediction.  Divergence analysis has confirmed that our synthetic data is indeed close to the test distribution.  Such augmented data is able to help validate models, which in turn can free up more recent data for training.  It also ensures that our models can be generalized to situations which have not happened before in this COVID-19 crisis and during some unprecedented moments, such as the wave caused by the variant Omicron which started in late 2021.

Our real-time results have been submitted to the US \cite{hub} and German \cite{hubde} Hubs weekly since Jun and Nov 2021, respectively.  They have contributed to these valuable datasets and also achieved some of the best results among all the submissions.  For references, our other retrospective results are also available in our forked repositories on GitHub \footnote{\url{https://github.com/cfong32/covid19-forecast-hub}} \footnote{\url{https://github.com/cfong32/covid19-forecast-hub-de}}.

\ifthenelse{\isundefined{\compilingmain}}{\subfile{sL_references}}{}

%% file: main.bbl
\begin{thebibliography}{10}
\providecommand{\url}[1]{#1}
\csname url@samestyle\endcsname
\providecommand{\newblock}{\relax}
\providecommand{\bibinfo}[2]{#2}
\providecommand{\BIBentrySTDinterwordspacing}{\spaceskip=0pt\relax}
\providecommand{\BIBentryALTinterwordstretchfactor}{4}
\providecommand{\BIBentryALTinterwordspacing}{\spaceskip=\fontdimen2\font plus
\BIBentryALTinterwordstretchfactor\fontdimen3\font minus
  \fontdimen4\font\relax}
\providecommand{\BIBforeignlanguage}[2]{{%
\expandafter\ifx\csname l@#1\endcsname\relax
\typeout{** WARNING: IEEEtran.bst: No hyphenation pattern has been}%
\typeout{** loaded for the language `#1'. Using the pattern for}%
\typeout{** the default language instead.}%
\else
\language=\csname l@#1\endcsname
\fi
#2}}
\providecommand{\BIBdecl}{\relax}
\BIBdecl

\bibitem{hub}
``{COVID}-19 {F}orecast {H}ub,'' 2020,
  \url{https://github.com/reichlab/covid19-forecast-hub}.

\bibitem{hubde}
``The {G}erman and {P}olish {COVID}-19 {F}orecast {H}ub,'' 2020,
  \url{https://github.com/KITmetricslab/covid19-forecast-hub-de}.

\bibitem{zou2020epidemic_ucla}
D.~Zou, L.~Wang, P.~Xu, J.~Chen, W.~Zhang, and Q.~Gu, ``Epidemic model guided
  machine learning for {COVID}-19 forecasts in the united states,''
  \emph{medRxiv}, 2020.

\bibitem{li2020forecasting_delphi}
M.~L. Li, H.~T. Bouardi, O.~S. Lami, T.~A. Trikalinos, N.~K. Trichakis, and
  D.~Bertsimas, ``Forecasting {COVID}-19 and analyzing the effect of government
  interventions,'' \emph{medRxiv}, 2020.

\bibitem{gibson2020real_MechBayes}
G.~C. Gibson, N.~G. Reich, and D.~Sheldon, ``Real-time mechanistic bayesian
  forecasts of covid-19 mortality,'' \emph{medRxiv}, 2020.

\bibitem{srivastava2020fast_alpha}
A.~Srivastava, T.~Xu, and V.~K. Prasanna, ``Fast and accurate forecasting of
  covid-19 deaths using the {SIkJ}$\alpha$ model,'' \emph{arXiv preprint
  arXiv:2007.05180}, 2020.

\bibitem{arik2020interpretable_google}
S.~Arik, C.-L. Li, J.~Yoon, R.~Sinha, A.~Epshteyn, L.~Le, V.~Menon, S.~Singh,
  L.~Zhang, M.~Nikoltchev \emph{et~al.}, ``Interpretable sequence learning for
  covid-19 forecasting,'' \emph{Advances in Neural Information Processing
  Systems}, vol.~33, 2020.

\bibitem{jia2020population_flow}
J.~S. Jia, X.~Lu, Y.~Yuan, G.~Xu, J.~Jia, and N.~A. Christakis, ``Population
  flow drives spatio-temporal distribution of {COVID}-19 in china,''
  \emph{Nature}, vol. 582, no. 7812, pp. 389--394, 2020.

\bibitem{marzouk2021deep_egypt}
M.~Marzouk, N.~Elshaboury, A.~Abdel-Latif, and S.~Azab, ``Deep learning model
  for forecasting covid-19 outbreak in egypt,'' \emph{Process Safety and
  Environmental Protection}, vol. 153, pp. 363--375, 2021.

\bibitem{yu2021covid_lstm}
C.-S. Yu, S.-S. Chang, T.-H. Chang, J.~L. Wu, Y.-J. Lin, H.-F. Chien, R.-J.
  Chen \emph{et~al.}, ``A covid-19 pandemic artificial intelligence--based
  system with deep learning forecasting and automatic statistical data
  acquisition: Development and implementation study,'' \emph{Journal of medical
  Internet research}, vol.~23, no.~5, p. e27806, 2021.

\bibitem{rodriguez2020deepcovid}
A.~Rodriguez, A.~Tabassum, J.~Cui, J.~Xie, J.~Ho, P.~Agarwal, B.~Adhikari, and
  B.~A. Prakash, ``{DeepCOVID}: An operational deep learning-driven framework
  for explainable real-time {COVID}-19 forecasting,'' \emph{medRxiv}, 2020.

\bibitem{jin2021interseries}
X.~Jin, Y.-X. Wang, and X.~Yan, ``Inter-series attention model for covid-19
  forecasting,'' in \emph{Proceedings of the 2021 SIAM International Conference
  on Data Mining (SDM)}.\hskip 1em plus 0.5em minus 0.4em\relax SIAM, 2021, pp.
  495--503.

\bibitem{callaway2021beyond_nature}
E.~Callaway \emph{et~al.}, ``Beyond omicron: what’s next for covid’s viral
  evolution,'' \emph{Nature}, vol. 600, no. 7888, pp. 204--207, 2021.

\bibitem{weiss2013sir}
H.~H. Weiss, ``The {SIR} model and the foundations of public health,''
  \emph{Materials matematics}, pp. 0001--17, 2013.

\bibitem{abbott2020estimating_delay}
S.~Abbott, J.~Hellewell, R.~N. Thompson, K.~Sherratt, H.~P. Gibbs, N.~I. Bosse,
  J.~D. Munday, S.~Meakin, E.~L. Doughty, J.~Y. Chun \emph{et~al.},
  ``Estimating the time-varying reproduction number of sars-cov-2 using
  national and subnational case counts,'' \emph{Wellcome Open Research},
  vol.~5, no. 112, p. 112, 2020.

\bibitem{chen2020state_seir}
S.~Chen, Q.~Li, S.~Gao, Y.~Kang, and X.~Shi, ``State-specific projection of
  covid-19 infection in the united states and evaluation of three major control
  measures,'' \emph{Scientific reports}, vol.~10, no.~1, pp. 1--9, 2020.

\bibitem{karlen2020characterizing}
D.~Karlen, ``Characterizing the spread of covid-19,'' \emph{arXiv preprint
  arXiv:2007.07156}, 2020.

\bibitem{wu2021deepgleam}
D.~Wu, L.~Gao, X.~Xiong, M.~Chinazzi, A.~Vespignani, Y.-A. Ma, and R.~Yu,
  ``Deepgleam: a hybrid mechanistic and deep learning model for covid-19
  forecasting,'' \emph{arXiv preprint arXiv:2102.06684}, 2021.

\bibitem{cramer2021evaluation_ensemble}
E.~Y. Cramer, V.~K. Lopez, J.~Niemi, G.~E. George, J.~C. Cegan, I.~D.
  Dettwiller, W.~P. England, M.~W. Farthing, R.~H. Hunter, B.~Lafferty
  \emph{et~al.}, ``Evaluation of individual and ensemble probabilistic
  forecasts of {COVID}-19 mortality in the us,'' \emph{medRxiv}, 2021.

\bibitem{bracher2021national_ensemble}
J.~Bracher, D.~Wolffram, J.~Deuschel, K.~Goergen, J.~L. Ketterer, A.~Ullrich,
  S.~Abbott, M.~V. Barbarossa, D.~Bertsimas, S.~Bhatia \emph{et~al.},
  ``National and subnational short-term forecasting of covid-19 in germany and
  poland, early 2021,'' \emph{medRxiv}, 2021.

\bibitem{salinas2020deepar}
D.~Salinas, V.~Flunkert, J.~Gasthaus, and T.~Januschowski, ``Deepar:
  Probabilistic forecasting with autoregressive recurrent networks,''
  \emph{International Journal of Forecasting}, vol.~36, no.~3, pp. 1181--1191,
  2020.

\bibitem{seo2018structured_GCRN}
Y.~Seo, M.~Defferrard, P.~Vandergheynst, and X.~Bresson, ``Structured sequence
  modeling with graph convolutional recurrent networks,'' in
  \emph{International Conference on Neural Information Processing}.\hskip 1em
  plus 0.5em minus 0.4em\relax Springer, 2018, pp. 362--373.

\bibitem{li2018diffusion_DCRNN}
Y.~Li, R.~Yu, C.~Shahabi, and Y.~Liu, ``Diffusion convolutional recurrent
  neural network: Data-driven traffic forecasting,'' in \emph{International
  Conference on Learning Representations}, 2018.

\bibitem{yu2018spatio_STGCN}
B.~Yu, H.~Yin, and Z.~Zhu, ``Spatio-temporal graph convolutional networks: a
  deep learning framework for traffic forecasting,'' in \emph{Proceedings of
  the 27th International Joint Conference on Artificial Intelligence}, 2018,
  pp. 3634--3640.

\bibitem{bai2020adaptive_AGCRN}
L.~BAI, L.~Yao, C.~Li, X.~Wang, and C.~Wang, ``Adaptive graph convolutional
  recurrent network for traffic forecasting,'' \emph{Advances in Neural
  Information Processing Systems}, vol.~33, 2020.

\bibitem{shang2021discrete_GTS}
C.~Shang, J.~Chen, and J.~Bi, ``Discrete graph structure learning for
  forecasting multiple time series,'' in \emph{International Conference on
  Learning Representations}, 2021.

\bibitem{vaswani2017attention}
A.~Vaswani, N.~Shazeer, N.~Parmar, J.~Uszkoreit, L.~Jones, A.~N. Gomez,
  {\L}.~Kaiser, and I.~Polosukhin, ``Attention is all you need,'' in
  \emph{Proceedings of the 31st International Conference on Neural Information
  Processing Systems}, 2017, pp. 6000--6010.

\bibitem{dosovitskiy2020image_ViT}
A.~Dosovitskiy, L.~Beyer, A.~Kolesnikov, D.~Weissenborn, X.~Zhai,
  T.~Unterthiner, M.~Dehghani, M.~Minderer, G.~Heigold, S.~Gelly \emph{et~al.},
  ``An image is worth 16x16 words: Transformers for image recognition at
  scale,'' in \emph{International Conference on Learning Representations},
  2020.

\bibitem{li2019enhancing}
S.~Li, X.~Jin, Y.~Xuan, X.~Zhou, W.~Chen, Y.-X. Wang, and X.~Yan, ``Enhancing
  the locality and breaking the memory bottleneck of transformer on time series
  forecasting,'' \emph{Advances in Neural Information Processing Systems},
  vol.~32, pp. 5243--5253, 2019.

\bibitem{zhou2020informer}
H.~Zhou, S.~Zhang, J.~Peng, S.~Zhang, J.~Li, H.~Xiong, and W.~Zhang,
  ``Informer: Beyond efficient transformer for long sequence time-series
  forecasting,'' in \emph{The Thirty-Fifth {AAAI} Conference on Artificial
  Intelligence, {AAAI} 2021}.\hskip 1em plus 0.5em minus 0.4em\relax {AAAI}
  Press, 2021, p. online.

\bibitem{tay2020long}
Y.~Tay, M.~Dehghani, S.~Abnar, Y.~Shen, D.~Bahri, P.~Pham, J.~Rao, L.~Yang,
  S.~Ruder, and D.~Metzler, ``Long range arena: A benchmark for efficient
  transformers,'' \emph{arXiv preprint arXiv:2011.04006}, 2020.

\bibitem{bochkovskiy2020yolov4}
A.~Bochkovskiy, C.-Y. Wang, and H.-Y.~M. Liao, ``Yolov4: Optimal speed and
  accuracy of object detection,'' \emph{arXiv preprint arXiv:2004.10934}, 2020.

\bibitem{wei2019eda}
J.~Wei and K.~Zou, ``Eda: Easy data augmentation techniques for boosting
  performance on text classification tasks,'' \emph{arXiv preprint
  arXiv:1901.11196}, 2019.

\bibitem{wen2020time_survey}
Q.~Wen, L.~Sun, F.~Yang, X.~Song, J.~Gao, X.~Wang, and H.~Xu, ``Time series
  data augmentation for deep learning: A survey,'' \emph{arXiv preprint
  arXiv:2002.12478}, 2020.

\bibitem{cramer2021united}
E.~Y. Cramer, Y.~Huang, Y.~Wang, E.~L. Ray, M.~Cornell, J.~Bracher, A.~Brennen,
  A.~J.~C. Rivadeneira, A.~Gerding, K.~House \emph{et~al.}, ``The united states
  covid-19 forecast hub dataset,'' \emph{medRxiv}, 2021.

\bibitem{bracher2021evaluating_wis}
J.~Bracher, E.~L. Ray, T.~Gneiting, and N.~G. Reich, ``Evaluating epidemic
  forecasts in an interval format,'' \emph{PLoS computational biology},
  vol.~17, no.~2, p. e1008618, 2021.

\bibitem{takeuchi2006nonparametric}
I.~Takeuchi, Q.~V. Le, T.~D. Sears, and A.~J. Smola, ``Nonparametric quantile
  estimation,'' \emph{Journal of Machine Learning Research}, vol.~7, pp.
  1231--1264, 2006.

\bibitem{wen2017multi}
R.~Wen, K.~Torkkola, B.~Narayanaswamy, and D.~Madeka, ``A multi-horizon
  quantile recurrent forecaster,'' \emph{arXiv preprint arXiv:1711.11053},
  2017.

\bibitem{rodrigues2020beyond}
F.~Rodrigues and F.~C. Pereira, ``Beyond expectation: Deep joint mean and
  quantile regression for spatiotemporal problems,'' \emph{IEEE transactions on
  neural networks and learning systems}, vol.~31, no.~12, pp. 5377--5389, 2020.

\bibitem{hansen2021assessment_reinfection}
C.~H. Hansen, D.~Michlmayr, S.~M. Gubbels, K.~M{\o}lbak, and S.~Ethelberg,
  ``Assessment of protection against reinfection with sars-cov-2 among 4
  million pcr-tested individuals in denmark in 2020: a population-level
  observational study,'' \emph{The Lancet}, vol. 397, no. 10280, pp.
  1204--1212, 2021.

\bibitem{byrd1995limited_LBFGSB}
R.~H. Byrd, P.~Lu, J.~Nocedal, and C.~Zhu, ``A limited memory algorithm for
  bound constrained optimization,'' \emph{SIAM Journal on scientific
  computing}, vol.~16, no.~5, pp. 1190--1208, 1995.

\bibitem{jhu}
``{COVID}-19 {D}ata {R}epository by the {C}enter for {S}ystems {S}cience and
  {E}ngineering ({CSSE}) at {J}ohns {H}opkins {U}niversity,'' 2020,
  \url{https://github.com/CSSEGISandData/COVID-19}.

\bibitem{van2008visualizing_tsne}
L.~Van~der Maaten and G.~Hinton, ``Visualizing data using t-sne.''
  \emph{Journal of machine learning research}, vol.~9, no.~11, 2008.

\bibitem{wang2009divergence_kNN}
Q.~Wang, S.~R. Kulkarni, and S.~Verd{\'u}, ``Divergence estimation for
  multidimensional densities via $ k $-nearest-neighbor distances,'' \emph{IEEE
  Transactions on Information Theory}, vol.~55, no.~5, pp. 2392--2405, 2009.

\bibitem{barbarossa2020impact_hubde_FIAS}
M.~V. Barbarossa, J.~Fuhrmann, J.~H. Meinke, S.~Krieg, H.~V. Varma,
  N.~Castelletti, and T.~Lippert, ``The impact of current and future control
  measures on the spread of covid-19 in germany,'' 2020.

\end{thebibliography}
